\documentclass{article}
\usepackage{iclr2016_workshop,times}
\usepackage{graphicx}
\usepackage{hyperref}
\usepackage{url}

\usepackage{amssymb}
\usepackage{bm}
\usepackage{tikz}
\usetikzlibrary{positioning,fit}

\title{Mixtures of Sparse Autoregressive Networks}

\author{Marc Goessling\\
Department of Statistics\\
University of Chicago\\
Chicago, IL 60637, USA\\
\texttt{goessling@galton.uchicago.edu}
\And Yali Amit\\
Departments of Statistics and Computer Science\\
University of Chicago\\
Chicago, IL 60637, USA\\
\texttt{amit@galton.uchicago.edu}}

\begin{document}

\maketitle

\begin{abstract}
We consider high-dimensional distribution estimation through autoregressive networks. By combining the concepts of sparsity, mixtures and parameter sharing we obtain a simple model which is fast to train and which achieves state-of-the-art or better results on several standard benchmark datasets. Specifically, we use an L1-penalty to regularize the conditional distributions and introduce a procedure for automatic parameter sharing between mixture components. Moreover, we propose a simple distributed representation which permits exact likelihood evaluations since the latent variables are interleaved with the observable variables and can be easily integrated out. Our model achieves excellent generalization performance and scales well to extremely high dimensions.
\end{abstract}

\section{Introduction}
We consider the fundamental task of learning a high-dimensional distribution $\mathbb{P}(\bm{x})$ from training examples $\bm{x}^{(n)}$. The recently most successful approaches are based on autoregressive networks which make use of the decomposition
\[
\mathbb{P}(\bm{x}) = \prod_{d=1}^D \mathbb{P}(x_d \,|\, \bm{x}_{1:d-1}).
\]
One simple way to model the $D$ conditional distributions is through logistic or linear regressions \citep{frey1998graphical}. Better results can be achieved by using flexible neural networks to model the conditionals \citep{bengio2000taking}. \citet{larochelle2011neural} proposed to use the same hidden units for all conditionals. Such a weight sharing among conditional distributions leads to a significant improvement in terms of the generalization performance \citep{bengio2011discussion,uria2013rnade}. Several variants of this idea have been proposed \citep{uria2014deep,raiko2014iterative,germain2015made}. Since the hidden units in these models are deterministic, the associated learning algorithms are relatively simple and exact likelihood evaluations are feasible. Neural networks with stochastic hidden units \citep{gregor2014deep,bornschein2015reweighted} on the other hand require sophisticated inference procedures and exact likelihood evaluations are then intractable.

In this work we start from sparse logistic and linear models for the conditionals. In contrast to nonsparse nonlinear models, these can be learned from relatively small datasets and have excellent generalization abilities. We then consider large mixtures of such sparse autoregressive networks (SpARN) and study the effect of parameter sharing among mixture components. Since there is no weight sharing between different conditional distributions, our learning procedure can be perfectly parallelized over the data dimensions. We are hence able to train models for extremely high-dimensional data. In our experiments we achieve state-of-the-art or better performance on several standard benchmark problems and obtain convincing results for datasets with several ten thousand dimensions.

\section{Sparse autoregressive networks}\label{sec:SpARN}
One major source of regularization in neural autoregressive networks is weight sharing among the conditional distributions for different dimensions. This seems necessary in order to avoid overfitting but makes it hard to parallelize the learning procedure. In addition, early stopping is commonly applied in neural networks to control the model complexity. Other typical regularization techniques are weight decay \citep{uria2013rnade,uria2014deep,raiko2014iterative}, dropout \citep{germain2015made} or adaptive weight noise \citep{gregor2014deep}. Weight decay means that the L2-norm of the model parameters is penalized.

None of these regularization methods yields sparse parameters. In this work we use an L1-penalty, which induces sparsity. For high-dimensional problems it is very reasonable to assume that only a small number of the variables $x_1,\ldots,x_{d-1}$ are relevant for predicting $x_d$. As shown in our experiments, the use of an L1-penalty yields much better results than an L2-penalty. We also tried elastic nets \citep{zou2005regularization}, which use an L1- as well as an L2-penalty, but the performance was typically worse than with just an L1-penalty. Since the effect of regularization depends on the scale of the variables, we encode binary data as $\pm1$ in order not to create a systematic bias. Similarly, for real-valued data we normalize the variables to have mean zero and variance one. The conditional distributions in our autoregressive network are modeled through L1-penalized logistic or linear regressions which are fitted separately for each dimension via coordinate descent.

%could use \left ... \right
Concretely, for binary data $\bm{x} \in \{-1,1\}^D$ we minimize
\[
\sum_{n=1}^N \log(1+\exp(-x_d^{(n)}[\alpha^{(d)}_0+(\bm{\alpha}^{(d)}_{1:d-1})^T\bm{x}_{1:d-1}^{(n)}])) + \lambda_0 |\alpha^{(d)}_0| + \lambda ||\bm{\alpha}^{(d)}_{1:d-1}||_1
\]
where $\lambda_0, \lambda > 0$. The intercept $\alpha^{(d)}_0$ has a special meaning and is hence regularized differently than the dependency weights $\alpha^{(d)}_i$, $i=1\,\ldots,d-1$. Shrinkage for the intercept is needed to avoid potentially degenerated probabilities (in case the empirical variance of $x_d$ is zero) but also improves the generalization performance. The penalty strength $\lambda_0$ can be much smaller than $\lambda$. The ratio $\lambda/\lambda_0$ is known as the intercept scaling factor. We do not expect the intercepts to be mostly zero, so a different type of penalty could be used. But for simplicity we penalize the absolute value.
%explicit intercept scaling allows to use a single penalty strength (e.g., liblinear)

For continuous data $\bm{x} \in \mathbb{R}^D$ we minimize
\[
\frac{1}{2}\sum_{n=1}^N (x^{(n)}_d - (\bm{\alpha}^{(d)}_{1:d-1})^T\bm{x}_{1:d-1}^{(n)} )^2 + \lambda ||\bm{\alpha}^{(d)}_{1:d-1}|||_1
\]
where $\lambda > 0$. The intercept is not needed here since all variables are centered in advance. The parameters $\bm{\alpha}^{(d)}_{1:d-1}$ specify the conditional mean $\mu^{(d)}(\bm{x}_{1:d-1})$ of $x_d$ given $\bm{x}_{1:d-1}$. The conditional standard deviation $\sigma^{(d)}$ is assumed to be fixed (i.e., it does not depend on $\bm{x}_{1:d-1}$) and is estimated from the residuals $x^{(n)}_d - \mu^{(d)}(\bm{x}_{1:d-1})$. If each conditional distribution $x_d | \bm{x}_{1:d-1}$ is Gaussian with a constant variance then $\bm{x}$ has a multivariate normal distribution. Alternatively, we could model the variance as a, say, quadratic function of the predicted value $\mu^{(d)}(\bm{x}_{1:d-1})$ in which case $\bm{x}$ would no longer be multivariate Gaussian. In any case, our conditional distributions are unimodal. We deal with multimodality by learning mixtures of autoregressive networks. In contrast to that, \citet{bishop1994mixture,davies2000mix} model the conditional distributions through mixtures but only learn a single network.

L1-penalized logistic and linear regressions for neighborhood selection have been used before in undirected graphical models \citep{meinshausen2006high,ravikumar2010high}. Sparse linear models for conditional distributions have been used in image analysis by \citet{cressie1998image,domke2008killed}. However, in these models the relevant parents for each pixel were chosen manually, consisting of a small number of adjacent pixels. This means that no long-range dependencies can be captured. Very restricted neighborhoods can lead to visible artifacts in samples form such models. A fully Bayesian approach for deep sigmoid belief nets with a sparsity-inducing prior has been used in \citet{gan2015learning}, where the posterior distribution on model parameters is approximated through a variational Bayes procedure. The quantitative performance of this method however is significantly worse than our L1-penalized regression approach (see Section \ref{sec:experiments}).

\subsection{Mixtures}
For large datasets a sparse autoregressive network is not competitive with more sophisticated networks. Since learning requires only relatively few training examples it is natural to consider mixtures of the form
\[
\mathbb{P}(\bm{x}) = \sum_{k=1}^K \mathbb{P}(h=k) \prod_{d=1}^D \mathbb{P}(x_d \,|\, \bm{x}_{1:d-1}, h)
\]
where the conditional distributions are sparse regressions. The latent variable $h \in \{1,\ldots,K\}$ indicates the active mixture component. It is straightforward to learn such a mixture using the EM-algorithm \citep{dempster1977maximum}. A good initialization can be obtained by first learning a mixture of product (Bernoulli or normal) distributions for the data.
%for small mixtures refitting can help, for large mixtures typically not since the assignment doesn't change

\subsubsection{Without parameter sharing}
In a classical mixture model no parameters are shared among the components. The form\footnote{We use here the notation for binary data, the corresponding expressions for continuous data are evident.} of the component conditionals in this case is
\[
\mathbb{P}(x_d=1 | \bm{x}_{1:d-1}, h) = \sigma(\beta_0^{(h,d)} + (\bm{\beta}^{(h,d)}_{1:d-1})^T\bm{x}_{1:d-1}).
\]
Consequently, separate autoregressive networks are trained for different clusters of the data. Without parameter sharing, a typical number of mixture components before overfitting occurs is around one per one thousand training examples.

\subsubsection{With parameter sharing}
A reasonable simplification is to assume that the dependence structure is rather universal, i.e., it does not change much for the different mixture components. With shared dependency weights a much larger mixture can be trained because only separate intercepts have to be learned. The form of the component conditionals with parameter sharing is
\[
\mathbb{P}(x_d=1 | \bm{x}_{1:d-1}, h) = \sigma(\beta_0^{(h,d)} + (\bm{\alpha}^{(d)}_{1:d-1})^T\bm{x}_{1:d-1}).
\]
The same sharing of dependency weights is used in autoregressive sigmoid belief networks \citep{gregor2014deep}, which are large implicit mixtures of (dense) autoregressive networks. Note that this kind of parameter sharing across mixture components is very different from the mentioned weight sharing in neural autoregressive networks \citep[e.g.,][]{larochelle2011neural} where weights are shared across conditionals for different dimensions. With shared dependency weights, a typical number of mixture components before overfitting occurs is around one per one hundred training examples. 

\subsubsection{Automatic parameter sharing}
Rather than manually deciding which parameters to share we can let the amount of sharing be part of the learning process. Specifically, we can use a global bias term and global dependency weights, and let the component parameters describe deviations from these global parameters. The form of the component conditionals in this case is
\[
\mathbb{P}(x_d=1 | \bm{x}_{1:d-1}, h) = \sigma(\alpha_0^{(d)} + \beta_0^{(h,d)} + (\bm{\alpha}^{(d)}_{1:d-1}+\bm{\beta}^{(h,d)}_{1:d-1})^T\bm{x}_{1:d-1}).
\]
We want the component parameters only to be nonzero if there is substantial gain from untying the corresponding weights. Consequently, we use the penalty
\[
\lambda_0 |\alpha^{(d)}_0| + \lambda_0 \sum_{h=1}^K |\beta^{(h,d)}_0| + \lambda ||\bm{\alpha}^{(d)}_{1:d-1}||_1 + \lambda \sum_{h=1}^K ||\bm{\beta}^{(h,d)}_{1:d-1}||_1.
\]
A similar penalty for deviations from a shared parameter has been used for multi-task learning with SVMs \citep{evgeniou2004regularized}. The optimal number of mixture components with automatic sharing is between the optimal numbers of components for mixtures with and without parameter sharing, respectively. In our experiments, mixtures with automatic parameter sharing always performed better than the mixtures with or without parameter sharing.

\subsection{Sequence of mixtures}
In a mixture model only a single component is active at a time. In contrast to that, a distributed representation \citep{bengio2013representation} allows several experts to cooperate in order to explain the observations. Each expert is a model for the data vector $\bm{x}$. Using a specified composition rule \citep{goessling2014compact} the active experts are combined to create the composed model for $\bm{x}$. The traditional approach is to use a top-down model in which latent variables are generated first and the distribution of the visible variables is modeled conditioned on all hidden variables. Such an organization of the variables makes inference and learning difficult. Moreover, exact likelihood evaluations are usually intractable because the required computations involve a sum over all configurations of the latent variables.
% requires approximate procedures

%The partition can in principle be arbitrarily, but it is best if the blocks consist of reusable/local information.
We propose a simple distributed representation for which inference is trivial and exact likelihood evaluations are tractable. The basic idea is to divide the data dimensions into chunks and to learn local models for each of these subsets of variables. Specifically, we partition the $D$ dimensions into $L$ disjoint intervals $I_\ell := [d_{\ell-1}+1:d_\ell]$ where
\[
0=d_0 < d_1 < \ldots < d_L = D.
\]
We then use mixtures of sparse autoregressive networks for each of the data vectors $\bm{x}_{I_\ell}$, $\ell=1,\ldots,L$, utilizing the data $\bm{x}_{1:d_{\ell-1}}$ from previous dimensions as additional predictors, i.e.,
\[
\mathbb{P}(\bm{x}_{I_\ell} | \bm{x}_{1:d_{\ell-1}},h_\ell) = \prod_{d=d_{\ell-1}+1}^{d_\ell} \mathbb{P}(x_d \,|\, \bm{x}_{1:d-1}, h_{\ell}).
\]

Associated with each mixture is a latent variable $h_\ell \in \{ 1,\ldots,K_\ell\}$ which specifies the component that is active for interval $I_\ell$. By choosing different configurations of mixture components global variation is achieved while the autoregressive networks themselves capture local variation of the data. For each choice of latent variables, the combined model is an autoregressive network for $\bm{x}$.

Partition-based representations are simpler and more interpretable than other distributed representations because the area of responsibility is explicitly defined for the local models. Moreover, since the local models describe disjoint sets of variables no composition rule is required (for each observable variable exactly one local model is employed). For image modeling the main drawbacks of existing partition-based approaches \citep{pal2002learning,aghajanian2008mosaicfaces} are discontinuities and misalignments along the partition boundaries. Since in our approach each local model is an autoregressive network that uses the data from all previous dimensions as predictors (not just the data within its own interval $I_\ell$) these problems are largely eliminated, see the experimental results in Section \ref{sec:experiments}.

%The latent and observable variables in the proposed architecture are interleaved in a linear order and the latent variables are conditionally independent given the observable variables

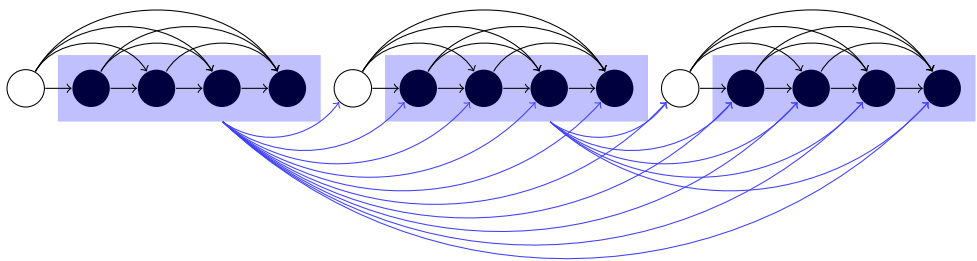
\begin{figure}
\center
\begin{tikzpicture}[scale=1.5,every node/.style={scale=1.5}]
\node (h1) [circle, draw=black]{};
\node (x1) [circle, fill=black, right=1em of h1]{};
\node (x2) [circle, fill=black, right=1em of x1]{};
\node (x3) [circle, fill=black, right=1em of x2]{};
\node (x4) [circle, fill=black, right=1em of x3]{};
\node (block1) [draw=blue!25, fit=(x1)(x4), fill=blue, fill opacity=0.25] {};

\draw[->]  (h1) to (x1);
\draw[bend left=60,->]  (h1) to (x2);
\draw[bend left=60,->]  (h1) to (x3);
\draw[bend left=60,->]  (h1) to (x4);
\draw[->]  (x1) to (x2);
\draw[bend left=60,->]  (x1) to (x3);
\draw[bend left=60,->]  (x1) to (x4);
\draw[->]  (x2) to (x3);
\draw[bend left=60,->]  (x2) to (x4);
\draw[->]  (x3) to (x4);

\node (h2) [circle, draw=black, right=1em of x4]{};
\node (x5) [circle, fill=black, right=1em of h2]{};
\node (x6) [circle, fill=black, right=1em of x5]{};
\node (x7) [circle, fill=black, right=1em of x6]{};
\node (x8) [circle, fill=black, right=1em of x7]{};
\node (block2) [draw=blue!25, fit= (x5)(x8), fill=blue, fill opacity=0.25] {};

\draw[bend right=45,blue!75,->]  (block1) to (h2);
\draw[bend right=45,blue!75,->]  (block1) to (x5);
\draw[bend right=45,blue!75,->]  (block1) to (x6);
\draw[bend right=45,blue!75,->]  (block1) to (x7);
\draw[bend right=45,blue!75,->]  (block1) to (x8);

\draw[->]  (h2) to (x5);
\draw[bend left=60,->]  (h2) to (x6);
\draw[bend left=60,->]  (h2) to (x7);
\draw[bend left=60,->]  (h2) to (x8);
\draw[->]  (x5) to (x6);
\draw[bend left=60,->]  (x5) to (x7);
\draw[bend left=60,->]  (x5) to (x8);
\draw[->]  (x6) to (x7);
\draw[bend left=60,->]  (x6) to (x8);
\draw[->]  (x7) to (x8);

\node (h3) [circle, draw=black, right=1em of x8]{};
\node (x9) [circle, fill=black, right=1em of h3]{};
\node (x10) [circle, fill=black, right=1em of x9]{};
\node (x11) [circle, fill=black, right=1em of x10]{};
\node (x12) [circle, fill=black, right=1em of x11]{};
\node (block3) [draw=blue!25, fit= (x9)(x12), fill=blue, fill opacity=0.25] {};

\draw[bend right=45,blue!75,->]  (block1) to (h3);
\draw[bend right=45,blue!75,->]  (block1) to (x9);
\draw[bend right=45,blue!75,->]  (block1) to (x10);
\draw[bend right=45,blue!75,->]  (block1) to (x11);
\draw[bend right=45,blue!75,->]  (block1) to (x12);

\draw[bend right=45,blue!75,->]  (block2) to (h3);
\draw[bend right=45,blue!75,->]  (block2) to (x9);
\draw[bend right=45,blue!75,->]  (block2) to (x10);
\draw[bend right=45,blue!75,->]  (block2) to (x11);
\draw[bend right=45,blue!75,->]  (block2) to (x12);

\draw[->]  (h3) to (x9);
\draw[bend left=60,->]  (h3) to (x10);
\draw[bend left=60,->]  (h3) to (x11);
\draw[bend left=60,->]  (h3) to (x12);
\draw[->]  (x9) to (x10);
\draw[bend left=60,->]  (x9) to (x11);
\draw[bend left=60,->]  (x9) to (x12);
\draw[->]  (x10) to (x11);
\draw[bend left=60,->]  (x10) to (x12);
\draw[->]  (x11) to (x12);
\end{tikzpicture}\vspace{-2em}
\caption{Graphical representation of the sequence-of-mixtures model. Solid nodes represent observable variables and empty nodes represent latent variables. Blue edges denote dependence on a block of variables.}
\label{fig:graphical_model}
\end{figure}

% context-sensitive
What remains to be specified is the distribution on the latent variables. In top-down approaches, for example deep autoregressive networks \citep{gregor2014deep}, the joint distribution on latent and observable variables is factored as $\mathbb{P}(\bm{h},\bm{x}) = \mathbb{P}(\bm{h}) \mathbb{P}(\bm{x} | \bm{h})$. This makes it easy to evaluate $\mathbb{P}(\bm{h})$ but hard to evaluate $\mathbb{P}(\bm{x})$. Rather than first generating all latent variables we propose to interleave them with the observable variables. Formally, the joint distribution is factored as
\[
\mathbb{P}(\bm{h},\bm{x}) = \prod_{\ell=1}^L \mathbb{P}(h_\ell|\bm{x}_{1:d_{\ell-1}}) \mathbb{P}(\bm{x}_{I_\ell} | \bm{x}_{1:d_{\ell-1}},h_\ell).
\]
where $\mathbb{P}(h_\ell|\bm{x}_{1:d_{\ell-1}})$ are multiclass logistic regressions. This means in particular that given $\bm{x}_{1:d_{\ell-1}}$ the distribution for $h_\ell$ does not depend on $\bm{h}_{1:\ell-1}$. Inference is thus trivial since the posterior distribution $\mathbb{P}(\bm{h} | \bm{x})$ factorizes over the latent variables. In contrast to that, hidden Markov models \citep{rabiner1989tutorial} define the distribution of $h_\ell$ in terms of $h_{\ell-1}$. Our factorization also makes it easy to integrate out the latent variables. Indeed, it follows that
\[
\mathbb{P}(\bm{x}) = \prod_{\ell=1}^L \sum_{h_\ell=1}^{K_\ell} \mathbb{P}(h_\ell|\bm{x}_{1:d_{\ell-1}}) \mathbb{P}(\bm{x}_{I_\ell} | \bm{x}_{1:d_{\ell-1}},h_\ell).
\]
Training the model hence reduces to learning $L$ separate mixtures of sparse autoregressive networks because the full likelihood $P(\bm{x})$ factorizes over $\ell$. Consequently, the model can be trained by learning $L$ separate mixtures of sparse autoregressive networks. A graphical representation of this sequence-of-mixtures model is shown in Figure \ref{fig:graphical_model}. The proposed model family is a subfamily of sum-product networks \citep{poon2011sum} which is particularly easy to train and conceptually simple.

\section{Experiments}\label{sec:experiments}
We quantitatively evaluated our sparse autoregressive networks (SpARN) on various high-dimensional datasets. For all experiments we used an intercept scaling factor (see Section \ref{sec:SpARN}) of 10, meaning the penalty on the intercept is a factor 10 smaller than for the dependency weights. The exact choice of the factor is not crucial, other values between 10 and 100 lead to very similar results. We used likelihood evaluations on the validation set to choose the number of mixture components and to select the penalty strength $\lambda$. The same penalty strength was applied for every dimension. We also performed a qualitative evaluation on two very high-dimensional datasets.

\subsection{Caltech-101 silhouettes}
The first dataset we considered are the Caltech-101 silhouettes \citep{marlin2010inductive}. There are 4,100 training samples, 2,264 validation samples and 2,307 test samples of dimension $28\times28 = 784$. Each data point is a binary mask for an object from one of 101 categories. Since the similarity between samples is of semantic nature a large degree of abstraction is required to generalize well to unseen examples. Previous results for this dataset are shown in Table \ref{tab:caltech101mnist}. The current state of the art is a logistic autoregressive network, also called a fully visible sigmoid belief net (FVSBN), which was learned through a variational Bayes procedure \citep{gan2015learning}. Our network based on L1-penalized logistic regressions and trained through coordinate descent achieves an average test log-likelihood which is 4.5 nats higher. An autoregressive network with L2-penalized logistic regressions performs significantly worse. The sparsity induced by the L1-penalty is thus crucial. We also trained mixtures of sparse autoregressive networks in an unsupervised manner (i.e., without using any category labels). The mixture with shared dependency weights (tied) improves over the single network by more than 2 nats. The optimal number of mixture components is 100, which makes sense because that is about the number of different object categories in the dataset. When trained without intercept scaling the mixture essentially reduced to a single sparse network. Hence, it was important to have a weaker penalty on the intercepts. The mixture with automatic parameter sharing (auto) further improves the test performance by about 1 nat. The mixture without parameter sharing (untied) was not better than the single network.

%FVSBN (VB) trained on only 10k examples?
\subsection{Binarized MNIST digits}
\begin{figure*}
\begin{center}
\includegraphics[width=.45\textwidth]{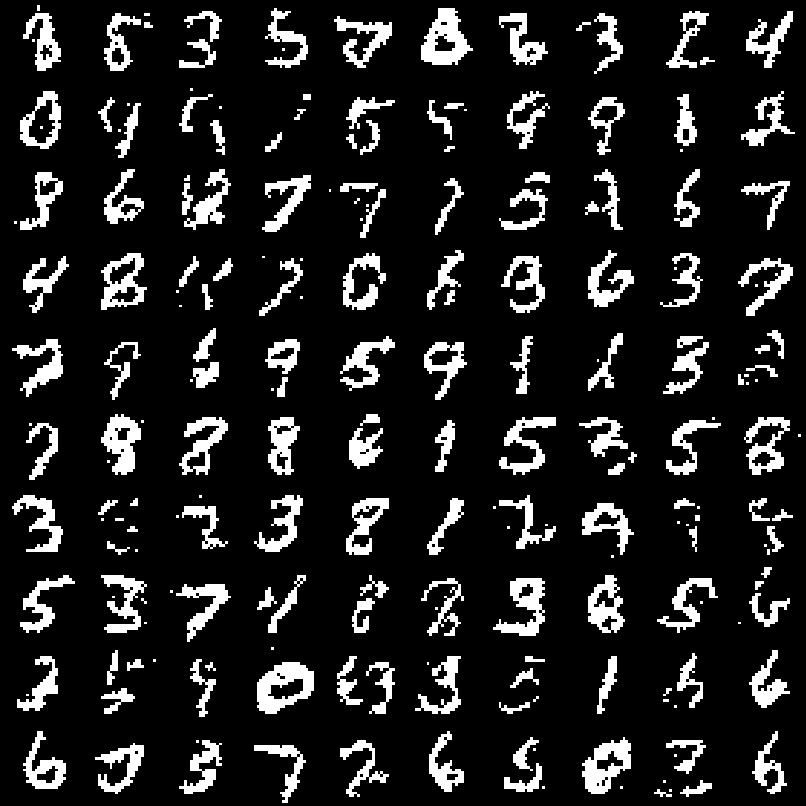}
\hspace{.04\textwidth}
\includegraphics[width=.45\textwidth]{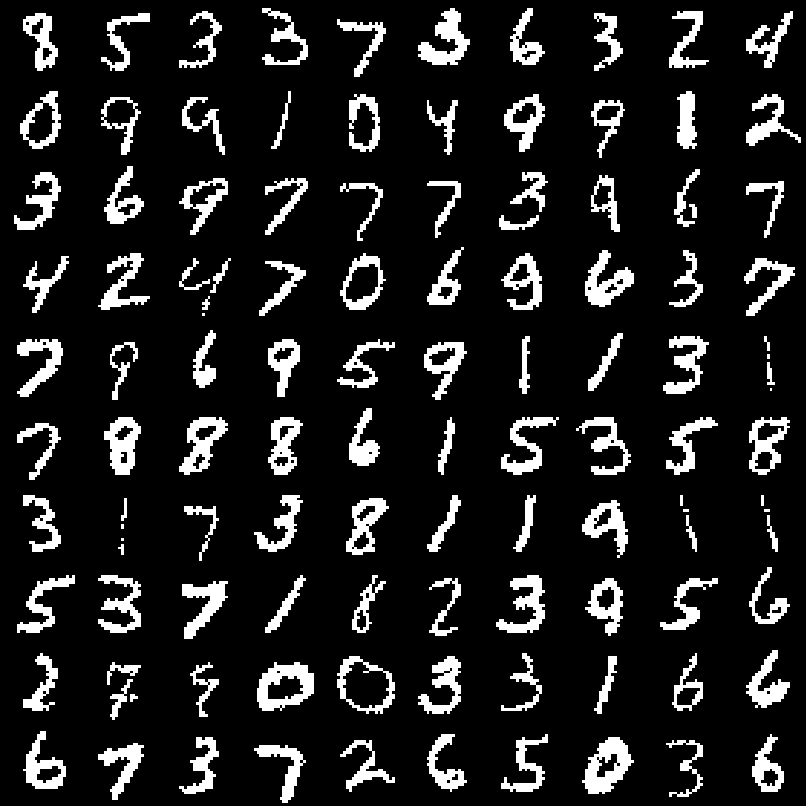}
\end{center}
\caption{\textbf{Left:} Samples from the sequence-of-mixtures model trained on MNIST digits. \textbf{Right:} Closest training examples.}
\label{fig:mnist}
\end{figure*}

The next dataset we considered are the MNIST digits \citep{salakhutdinov2008quantitative} which were binarized through random sampling based on the original intensity value. This is a relatively large dataset consisting of 50,000 training samples, 10,000 validation samples and 10,000 test samples again of dimension $28\times28 = 784$. Previous results for this dataset are shown in Table \ref{tab:caltech101mnist}. A fully visible sigmoid belief net trained by stochastic gradient descent and regularized through early stopping achieves about the same performance as a sparse logistic autoregressive network. This is because the dataset is rather large and hence the need for regularization is small.For mixtures without parameter sharing the optimal number of components is 20. Using automatic parameter sharing it is 50 and with shared dependency weights it is 500. All mixtures substantially improve over the single network. For this dataset, the mixture without sharing performs better than the mixture with shared parameters. We believe that this is the case because compared to the Caltech-101 silhouettes the distribution of MNIST digits has fewer modes and more training samples are available to learn from. We also trained a sequence-of-mixtures model. For that the image grid was divided into four quadrants of size $14\times14$ pixels and a mixture model was learned for each of the quadrants using the same number of components as before. With parameter sharing this model improves the performance of the corresponding mixture by 2.5 nats. Note that the number of parameters for the sequence-of-mixtures model is almost the same as for the mixture. The only additional parameters come from the multiclass logistic regressions for the hidden variables. When automatic parameter sharing is used the gain through a sequence of mixtures is smaller. Without parameter sharing there is no gain from the sequence of mixtures. Samples from the sequence-of-mixtures model with automatic parameter sharing as well as nearest training examples are shown in Figure \ref{fig:mnist}. This shows that the model puts the probability mass in the right region of the data space and that it does not simply memorize the training examples.

%Ber Mix results on those?
\begin{table*} %[b]
\caption{Average log-likelihoods (in nats) per test example using different models. The best result for each dataset is shown in bold. Baseline results are taken from [1]\citet{raiko2014iterative}, [2]\citet{bornschein2015reweighted}, [3]\citet{gan2015learning}, [4]\citet{larochelle2011neural}, [5]\citet{germain2015made} and [6]\citet{gregor2014deep}.}
\label{tab:caltech101mnist}
\begin{center}
\scriptsize
\begin{tabular}{lr}
\hline
\multicolumn{2}{c}{\bf Caltech-101 silhouettes}\\
\hline
RBM [1] & $\approx$-107.78\\
NADE-5 [1] & -107.28\\
RWS-NADE [2] & $\approx$-104.30\\
FVSBN (VB) [3]& $\approx$-96.40\\
\hline
ARN (1 comp, L2-penalty) & -95.95\\
SpARN (1 comp) & -91.80\\
SpARN (100 comp, auto) & \bf -88.48\\
SpARN (100 comp, tied) & -89.59
\end{tabular}
\hspace{3em}
\begin{tabular}{lr|lr}
\hline
\multicolumn{4}{c}{\bf Binarized MNIST digits}\\
\hline
FVSBN (VB) [3] & $\approx$-100.76 & SpARN (1 comp) & -97.34\\
FVSBN (SGD) [4] & -97.45 & SpARN (20 comp, untied) & -89.43\\
NADE [4] & -88.86 & SpARN (50 comp, auto) & -87.95\\
MADE [5] & -86.64 & SpARN (500 comp, tied) & -91.32\\
RBM (CD-25) [4] & $\approx$-86.34 & SpARN (4$\times$50 comp, auto) & -87.40\\
DARN [6] & $\approx$ \bf -84.13 & SpARN (4$\times$500 comp, tied) & -88.63
\end{tabular}
\end{center}
\end{table*}

%\begin{figure*}[h!tb]
%\includegraphics[width=\textwidth]{silhouettes_samples2}
%
%\vspace{1em}
%
%\includegraphics[width=\textwidth]{silhouettes_nearest2}
%\end{figure*}

\subsection{Binary UCI datasets}
We performed additional quantitative evaluations on eight standard benchmark datasets from the UCI Machine Learning Repository using the same splits into training, validation and test sets as in previous work. The datasets have various sample sizes and dimensions. The ratio of training samples to dimensionality varies between 1 and 300. Table \ref{tab:binaryuci} reports the obtained test log-likelihoods for mixtures of sparse autoregressive networks together with previous results. The table also shows the optimal numbers of mixture components for the different kinds of parameter sharing and the standard error of the test log-likelihood (when using automatic parameter sharing). On smaller datasets mixtures with parameter sharing tend to perform a bit better than mixtures without parameter sharing, and vice versa on larger datasets. The performance of our mixtures of sparse autoregressive networks with automatic parameter sharing is better than the best previously reported result on six of the eight datasets (all improvements except for the NIPS-0-12 dataset are statistically significant). For the other two datasets the performance is close to the state of the art.
%conceputally simple and easy to train

\begin{table*}
\caption{Average log-likelihoods (in nats) per test example for various models and datasets. The best result for each dataset is shown in bold. Baseline results are taken from \citet{germain2015made,bornschein2015reweighted}. Dataset sizes, standard errors and used numbers of mixture components are shown in italic.}
\label{tab:binaryuci}
\begin{center}
\scriptsize
\begin{tabular}{l*{9}{c}}
\hline
& \bf Adult & \bf Connect4 & \bf DNA & \bf Mushrooms & \bf NIPS-0-12 & \bf OCR-letters & \bf RCV1 & \bf Web\\
\hline
\em train & \em 5,000 & \em 16,000 & \em 1,400 & \em 2,000 & \em 400 & \em 32,152 & \em 40,000 & \em 14,000\\
\em valid & \em 1,414 & \em 4,000 & \em 600 & \em 500 & \em 100 & \em 10,000 & \em 10,000 & \em 3,188\\
\em test & \em 26,147 & \em 47,557 & \em 1,186 & \em 5,624 & \em 1,240 & \em 10,000 & \em 150,000 & \em 32,561\\
\em dim & \em 123 & \em 126 & \em 180 & \em 112 & \em 500 & \em 128 & \em 150 & \em 300\\
~\\
Ber. Mix. & -20.44 & -23.41 & -98.19 & -14.46 & -290.02 & -40.56 & -47.59 & 30.16\\
FVSBN (SGD) & -13.17 & -12.39 & -83.64 & -10.27 & -276.88 & -39.30 & -49.84 & -29.35\\
NADE & -13.19 & -11.99 & -84.81 & -9.81 & -273.08 & -27.22 & -46.66 & -28.39\\
DARN & -13.19 & -11.91 & -81.04 & -9.55 & -274.68 & $\approx$-28.17 & $\approx$-46.10 & $\approx$-28.83\\
MADE & -13.12 & -11.90 & -79.66 & -9.68 & -277.28 & -28.34 & -46.74 & -28.25\\
RWS-NADE & $\approx$-13.16 & $\approx$\bf-11.68 & $\approx$-84.26 & $\approx$-9.71 & $\approx$-271.11 & $\approx$\bf-26.43 & $\approx$-46.09 & $\approx$-27.92\\
\hline
SpARN (untied) & \bf-13.04 & -12.04 & -79.32 & -9.58 & -271.13 & -28.48 & -45.55 & -27.98\\
SpARN (auto) & \bf-13.04 & -11.98 & \bf-79.05 & \bf-9.38 & \bf-270.29 & -27.96 & \bf-45.11 & \bf-27.50\\
SpARN (tied) & \bf-13.04 & -12.14 & -79.26 & -9.39 & -270.53 & -31.53 & -45.50 & -28.12\\
~\\
\em std. error & \em 0.02 & \em 0.01 & \em 0.21 & \em 0.01 & \em 0.52 & \em 0.11 & \em 0.06 & \em 0.10\\
\em comp (untied) & \em 1 & \em 10 & \em 1 & \em 2 & \em 1 & \em 100 & \em 200 & \em 100\\
\em comp (auto) & \em 1 & \em 10 & \em 3 & \em 10 & \em 20 & \em 200 & \em 1,000 & \em 200\\
\em comp (tied) & \em 1 & \em 10 & \em 3 & \em 20 & \em 20 & \em 5,000 & \em 5,000 & \em 200
\end{tabular}
\end{center}
\end{table*}

\begin{figure*}[b!]
\includegraphics[width=\textwidth]{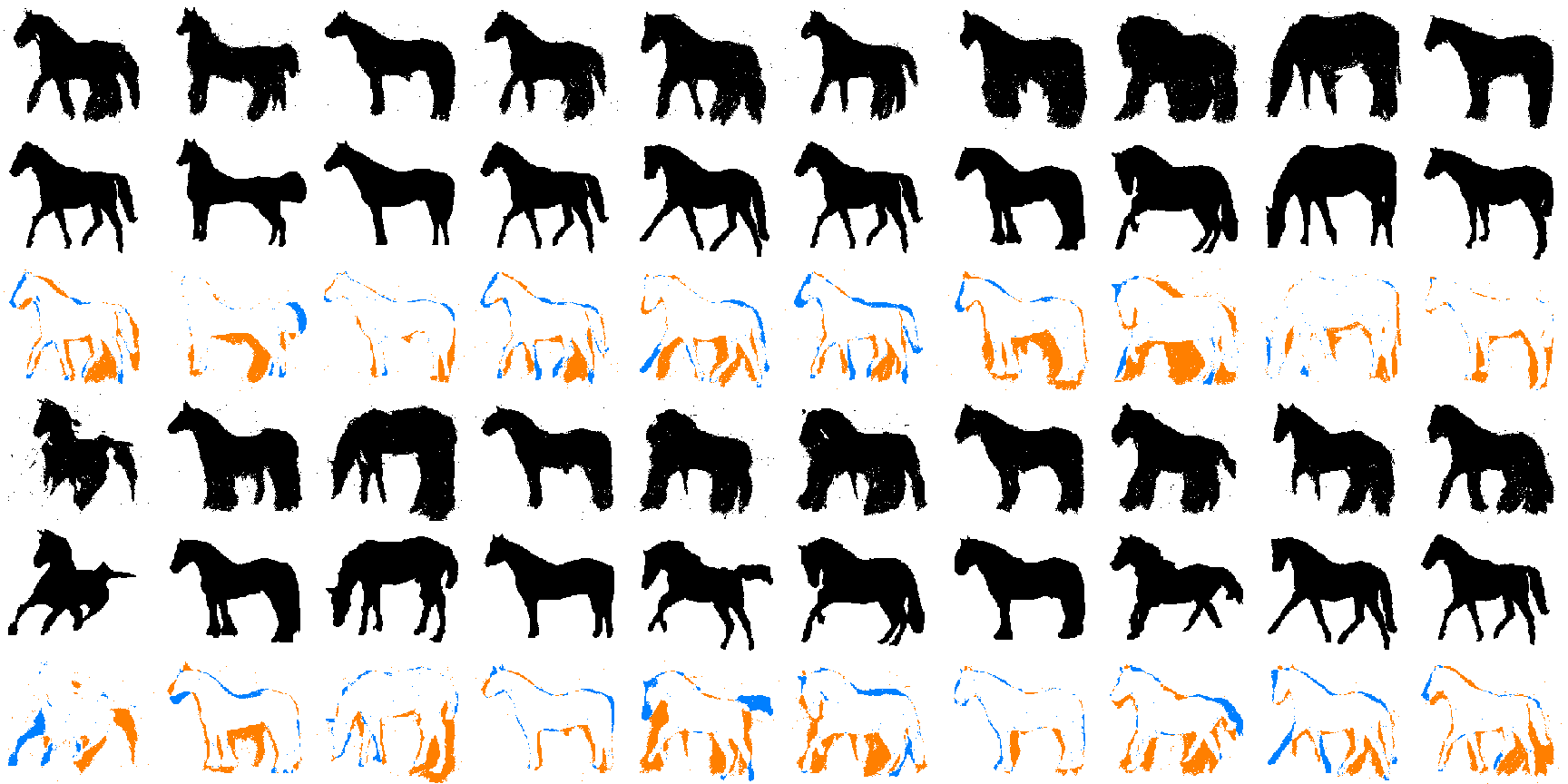}
\caption{\textbf{1st \& 4th row:} Samples from the sequence-of-mixtures model trained on Weizmann horses. \textbf{2nd \& 5th row:} Closest training examples. \textbf{3rd \& 6th row:} Symmetric differences between the synthetic sample and the closest training example (pixels in blue are only `on' for the training example and pixels in orange are only `on' for the synthetic sample).}
\label{fig:horses}
\end{figure*}

\subsection{Weizmann horses}
The following experiment shows the ability of our model to cope with very high-dimensional data. The Weizmann horse dataset \citep{borenstein2008combined} consists of 328 binary images of size 200$\times$240. We divided the image grid into four quadrants of size 100$\times$120 and learned a mixture of sparse autoregressive networks for each quadrant. We decided to use five components each for the top two quadrants (i.e., horse head and back) and ten components each for the bottom two quadrants (i.e., horse legs) because there is more variability in the lower half of the image. Samples from the sequence-of-mixtures model are presented in Figure \ref{fig:horses} together with the closest training examples. Our model creates overall realistic samples which differ from the training examples. In particular, no discontinuities along the borders of the quadrants are visible. We emphasize that our model was trained using the full resolution of the images. Each data point is a 48,000 dimensional binary vector. The same dataset was used in \citet{eslami2014shape} where a Boltzmann machine was learned from the data. In their experiments the images were downsampled to $32\times32$ pixels, which is a factor 50 smaller than in our case. This was probably necessary in order to speed up computations and to avoid overfitting. Refer to Figure 5(d) in that paper to compare the quality of the model samples.

\begin{figure*}[b!]
\includegraphics[width=\textwidth]{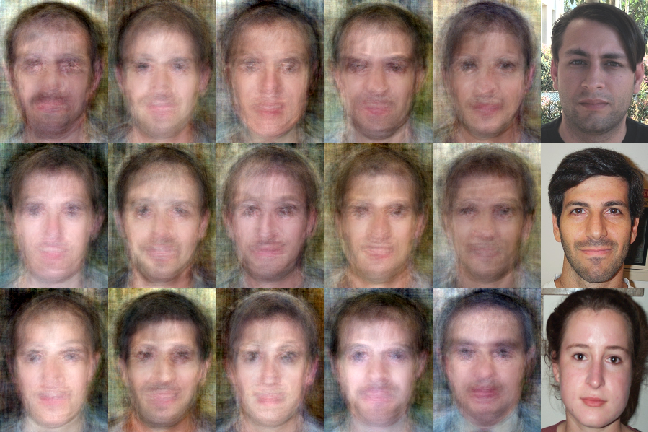}
\caption{\textbf{First five columns:} Samples from a sparse linear autoregressive network trained on Caltech faces. \textbf{Last column:} Closest training examples for the adjacent sample.}
\label{fig:faces}
\end{figure*}

\subsection{Caltech faces}
A very high-dimensional real-valued dataset that we considered are the Caltech Faces \citep{fergus2003object} consisting of 450 high-resolution color images. We downsampled the images to $200\times150\times3$, which are 90,000 dimensions. Samples from a sparse linear autoregressive network with constant conditional variance are presented in Figure \ref{fig:faces}. The samples are diverse and show substantial abstraction from the training examples. We also learned a network in which the conditional variance was modeled as a quadratic function of the predicted value. This improved the likelihood only slightly and the corresponding samples were visually indistinguishable from the samples of the network with constant variance. Since the variability in this dataset is mainly local, a mixture of sparse autoregressive networks did not improve over the single network in terms of likelihoods.
% The nearest neighbors indeed correspond to the perceptually closest training samples (based on manual inspection).

\section{Conclusion}
%possible improvements: different regularization per dimension (i.e., choice of lam)
We pointed out the importance of sparsity in autoregressive networks. Using simple mixtures of sparse autoregressive networks we achieved results which are competitive with much more sophisticated models. We also introduced a distributed representation for autoregressive networks based on a sequence of mixture models. No weight sharing across dimensions was needed for our models which allowed us to perfectly parallelize the learning procedure. %Python code for our model will be made publicly available.
% i.e., the training time decreases linearly with the number of cores (if there are as many cores as there are dimensions then learning our model only takes as long as fitting a single logistic/linear regression).

%\subsubsection*{Acknowledgments}

\bibliography{iclr2016_paper}

\begin{thebibliography}{31}
\providecommand{\natexlab}[1]{#1}
\providecommand{\url}[1]{\texttt{#1}}
\expandafter\ifx\csname urlstyle\endcsname\relax
  \providecommand{\doi}[1]{doi: #1}\else
  \providecommand{\doi}{doi: \begingroup \urlstyle{rm}\Url}\fi

\bibitem[Aghajanian \& Prince(2008)Aghajanian and
  Prince]{aghajanian2008mosaicfaces}
Jania Aghajanian and Simon~JD Prince.
\newblock Mosaicfaces: a discrete representation for face recognition.
\newblock In \emph{IEEE Workshop on Applications of Computer Vision}, pp.\
  1--8, 2008.

\bibitem[Bengio \& Bengio(2000)Bengio and Bengio]{bengio2000taking}
Samy Bengio and Yoshua Bengio.
\newblock Taking on the curse of dimensionality in joint distributions using
  neural networks.
\newblock \emph{IEEE Transactions on Neural Networks}, 11\penalty0
  (3):\penalty0 550--557, 2000.

\bibitem[Bengio(2011)]{bengio2011discussion}
Yoshua Bengio.
\newblock Discussion of the neural autoregressive distribution estimator.
\newblock In \emph{International Conference on Artificial Intelligence and
  Statistics}, pp.\  38--39, 2011.

\bibitem[Bengio et~al.(2013)Bengio, Courville, and
  Vincent]{bengio2013representation}
Yoshua Bengio, Aaron Courville, and Pierre Vincent.
\newblock Representation learning: A review and new perspectives.
\newblock \emph{IEEE Transactions on Pattern Analysis and Machine
  Intelligence}, 35\penalty0 (8):\penalty0 1798--1828, 2013.

\bibitem[Bishop(1994)]{bishop1994mixture}
Christopher~M Bishop.
\newblock Mixture density networks.
\newblock Technical Report NCRG/94/004, Aston University, Birmingham, 1994.

\bibitem[Borenstein \& Ullman(2008)Borenstein and
  Ullman]{borenstein2008combined}
Eran Borenstein and Shimon Ullman.
\newblock Combined top-down/bottom-up segmentation.
\newblock \emph{IEEE Transactions on Pattern Analysis and Machine
  Intelligence}, 30\penalty0 (12):\penalty0 2109--2125, 2008.

\bibitem[Bornschein \& Bengio(2015)Bornschein and
  Bengio]{bornschein2015reweighted}
J{\"o}rg Bornschein and Yoshua Bengio.
\newblock Reweighted wake-sleep.
\newblock In \emph{International Conference on Learning Representations}, 2015.

\bibitem[Cressie \& Davidson(1998)Cressie and Davidson]{cressie1998image}
Noel Cressie and Jennifer~L Davidson.
\newblock Image analysis with partially ordered markov models.
\newblock \emph{Computational Statistics \& Data Analysis}, 29\penalty0
  (1):\penalty0 1--26, 1998.

\bibitem[Davies \& Moore(2000)Davies and Moore]{davies2000mix}
Scott Davies and Andrew Moore.
\newblock Mix-nets: Factored mixtures of gaussians in bayesian networks with
  mixed continuous and discrete variables.
\newblock In \emph{Conference on Uncertainty in Artificial Intelligence}, pp.\
  168--175, 2000.

\bibitem[Dempster et~al.(1977)Dempster, Laird, and Rubin]{dempster1977maximum}
Arthur~P Dempster, Nan~M Laird, and Donald~B Rubin.
\newblock Maximum likelihood from incomplete data via the em algorithm.
\newblock \emph{Journal of the Royal Statistical Society. Series B
  (methodological)}, pp.\  1--38, 1977.

\bibitem[Domke et~al.(2008)Domke, Karapurkar, and Aloimonos]{domke2008killed}
Justin Domke, Alap Karapurkar, and Yiannis Aloimonos.
\newblock Who killed the directed model?
\newblock In \emph{Conference on Computer Vision and Pattern Recognition}, pp.\
   1--8, 2008.

\bibitem[Eslami et~al.(2014)Eslami, Heess, Williams, and Winn]{eslami2014shape}
SM~Ali Eslami, Nicolas Heess, Christopher~KI Williams, and John Winn.
\newblock The shape boltzmann machine: a strong model of object shape.
\newblock \emph{International Journal of Computer Vision}, 107\penalty0
  (2):\penalty0 155--176, 2014.

\bibitem[Evgeniou \& Pontil(2004)Evgeniou and Pontil]{evgeniou2004regularized}
Theodoros Evgeniou and Massimiliano Pontil.
\newblock Regularized multi--task learning.
\newblock In \emph{International conference on knowledge discovery and data
  mining}, pp.\  109--117, 2004.

\bibitem[Fergus et~al.(2003)Fergus, Perona, and Zisserman]{fergus2003object}
Robert Fergus, Pietro Perona, and Andrew Zisserman.
\newblock Object class recognition by unsupervised scale-invariant learning.
\newblock In \emph{Conference on Computer Vision and Pattern Recognition},
  volume~2, pp.\  264--271, 2003.

\bibitem[Frey(1998)]{frey1998graphical}
Brendan~J Frey.
\newblock \emph{Graphical models for machine learning and digital
  communication}.
\newblock MIT press, 1998.

\bibitem[Gan et~al.(2015)Gan, Henao, Carlson, and Carin]{gan2015learning}
Zhe Gan, Ricardo Henao, David Carlson, and Lawrence Carin.
\newblock Learning deep sigmoid belief networks with data augmentation.
\newblock In \emph{International Conference on Artificial Intelligence and
  Statistics}, pp.\  268--276, 2015.

\bibitem[Germain et~al.(2015)Germain, Gregor, Murray, and
  Larochelle]{germain2015made}
Mathieu Germain, Karol Gregor, Iain Murray, and Hugo Larochelle.
\newblock Made: masked autoencoder for distribution estimation.
\newblock In \emph{International Conference on Machine Learning}, pp.\
  881--889, 2015.

\bibitem[Goessling \& Amit(2015)Goessling and Amit]{goessling2014compact}
Marc Goessling and Yali Amit.
\newblock Compact part-based image representations.
\newblock In \emph{International Conference on Learning Representations
  (Workshop)}, 2015.

\bibitem[Gregor et~al.(2014)Gregor, Danihelka, Mnih, Blundell, and
  Wierstra]{gregor2014deep}
Karol Gregor, Ivo Danihelka, Andriy Mnih, Charles Blundell, and Daan Wierstra.
\newblock Deep autoregressive networks.
\newblock In \emph{International Conference on Machine Learning}, pp.\
  1242--1250, 2014.

\bibitem[Larochelle \& Murray(2011)Larochelle and Murray]{larochelle2011neural}
Hugo Larochelle and Iain Murray.
\newblock The neural autoregressive distribution estimator.
\newblock In \emph{International Conference on Artificial Intelligence and
  Statistics}, pp.\  29--37, 2011.

\bibitem[Marlin et~al.(2010)Marlin, Swersky, Chen, and
  Freitas]{marlin2010inductive}
Benjamin~M Marlin, Kevin Swersky, Bo~Chen, and Nando~D Freitas.
\newblock Inductive principles for restricted boltzmann machine learning.
\newblock In \emph{International Conference on Artificial Intelligence and
  Statistics}, pp.\  509--516, 2010.

\bibitem[Meinshausen \& B{\"u}hlmann(2006)Meinshausen and
  B{\"u}hlmann]{meinshausen2006high}
Nicolai Meinshausen and Peter B{\"u}hlmann.
\newblock High-dimensional graphs and variable selection with the lasso.
\newblock \emph{The Annals of Statistics}, pp.\  1436--1462, 2006.

\bibitem[Pal et~al.(2002)Pal, Frey, and Jojic]{pal2002learning}
Chris Pal, Brendan~J Frey, and Nebojsa Jojic.
\newblock Learning montages of transformed latent images as representations of
  objects that change in appearance.
\newblock In \emph{Computer Vision -- ECCV}, pp.\  715--731. 2002.

\bibitem[Poon \& Domingos(2011)Poon and Domingos]{poon2011sum}
Hoifung Poon and Pedro Domingos.
\newblock Sum-product networks: A new deep architecture.
\newblock In \emph{International Conference on Computer Vision (Workshops)},
  pp.\  689--690, 2011.

\bibitem[Rabiner(1989)]{rabiner1989tutorial}
Lawrence~R Rabiner.
\newblock A tutorial on hidden markov models and selected applications in
  speech recognition.
\newblock \emph{Proceedings of the IEEE}, 77\penalty0 (2):\penalty0 257--286,
  1989.

\bibitem[Raiko et~al.(2014)Raiko, Li, Cho, and Bengio]{raiko2014iterative}
Tapani Raiko, Yao Li, Kyunghyun Cho, and Yoshua Bengio.
\newblock Iterative neural autoregressive distribution estimator nade-k.
\newblock In \emph{Advances in Neural Information Processing Systems}, pp.\
  325--333, 2014.

\bibitem[Ravikumar et~al.(2010)Ravikumar, Wainwright, and
  Lafferty]{ravikumar2010high}
Pradeep Ravikumar, Martin~J Wainwright, and John~D Lafferty.
\newblock High-dimensional ising model selection using $\ell$1-regularized
  logistic regression.
\newblock \emph{The Annals of Statistics}, 38\penalty0 (3):\penalty0
  1287--1319, 2010.

\bibitem[Salakhutdinov \& Murray(2008)Salakhutdinov and
  Murray]{salakhutdinov2008quantitative}
Ruslan Salakhutdinov and Iain Murray.
\newblock On the quantitative analysis of deep belief networks.
\newblock In \emph{International Conference on Machine learning}, pp.\
  872--879, 2008.

\bibitem[Uria et~al.(2013)Uria, Murray, and Larochelle]{uria2013rnade}
Benigno Uria, Iain Murray, and Hugo Larochelle.
\newblock Rnade: The real-valued neural autoregressive density-estimator.
\newblock In \emph{Advances in Neural Information Processing Systems}, pp.\
  2175--2183, 2013.

\bibitem[Uria et~al.(2014)Uria, Murray, and Larochelle]{uria2014deep}
Benigno Uria, Iain Murray, and Hugo Larochelle.
\newblock A deep and tractable density estimator.
\newblock In \emph{International Conference on Machine Learning}, pp.\
  467--475, 2014.

\bibitem[Zou \& Hastie(2005)Zou and Hastie]{zou2005regularization}
Hui Zou and Trevor Hastie.
\newblock Regularization and variable selection via the elastic net.
\newblock \emph{Journal of the Royal Statistical Society: Series B (Statistical
  Methodology)}, 67\penalty0 (2):\penalty0 301--320, 2005.

\end{thebibliography}
\bibliographystyle{iclr2016_workshop}

\end{document}